\documentclass[letterpaper, 10 pt, conference]{ieeeconf}  

\IEEEoverridecommandlockouts                              
\usepackage{cite}
\usepackage{amsmath,amssymb,amsfonts}
\usepackage{algorithmic}
\usepackage{graphicx}
\usepackage{textcomp}
\usepackage{booktabs} 
\usepackage{xcolor}
\usepackage{hyperref}
\title{\LARGE \bf
Depth Restoration of Hand-Held Transparent Objects for Human-to-Robot Handover
}

\author{Ran Yu$^{1*}$, Haixin Yu$^{1*}$, Shoujie Li$^{1*}$, Yan Huang$^{1}$, Ziwu Song$^{1}$, Wenbo Ding$^{1,2\dag}$
\thanks{*Authors with equal contributions. \dag Corresponding author.}
\thanks{$^{1}$Shenzhen Ubiquitous Data Enabling Key Lab, Shenzhen International Graduate School, Tsinghua University, Shenzhen 518055, China. (Emails: yur23@mails.tsinghua.edu.cn, yuhx21@tsinghua.org.cn, \{lsj20, Huang-y24, song-zw20\}@mails.tsinghua.edu.cn and ding.wenbo@sz.tsinghua.edu.cn)}%
\thanks{$^{2}$RISC-V International Open Source Laboratory,Shenzhen 518055,China.}
\thanks{This work was supported by Shenzhen Key Laboratory of Ubiquitous Data Enabling (No. ZDSYS20220527171406015), Shenzhen Science and Technology Program (JCYJ20220530143013030), Guangdong Innovative and Entrepreneurial Research Team Program (2021ZT09L197), Tsinghua Shenzhen International Graduate School-Shenzhen Pengrui Young Faculty Program of Shenzhen Pengrui Foundation (No. SZPR2023005), Shenzhen Higher Education Stable Support Program  (WDZC20231129093657002), and Meituan.}
\thanks{Project website: \href{https://marcyu0303.github.io/HADR.github.io/}{\textcolor{blue}{https://marcyu0303.github.io/HADR.github.io/}}.}
}
\begin{document}

\maketitle
\thispagestyle{empty}
\pagestyle{empty}

\begin{abstract}

Transparent objects are common in daily life, while their optical properties pose challenges for RGB-D cameras to capture accurate depth information. This issue is further amplified when these objects are hand-held,  as hand occlusions further complicate depth estimation.  For assistant robots, however, accurately perceiving hand-held transparent objects is critical to effective human-robot interaction. This paper presents a Hand-Aware Depth Restoration (HADR) method based on creating an implicit neural representation function from a single RGB-D image. The proposed method utilizes hand posture as an important guidance to leverage semantic and geometric information of hand-object interaction. To train and evaluate the proposed method, we create a high-fidelity synthetic dataset named TransHand-14K with a real-to-sim data generation scheme. Experiments show that our method has better performance and generalization ability compared with existing methods. We further develop a real-world human-to-robot handover system based on HADR, demonstrating its potential in human-robot interaction applications.

\end{abstract}

\section{Introduction}
RGB-D cameras have become popular for robotic perception due to their ability to capture 3D information, which is crucial for robot grasping~\cite{mousavian20196} and manipulation~\cite{ze20243d}. However, transparent objects pose challenges in capturing correct depth information from RGB-D cameras due to light refraction and reflection~\cite{sajjan2020clear}. This issue is particularly critical for service robots~\cite{jiang2023robotic}, which are deployed in scenarios where transparent objects widely exist. Given the growing demand for human-robot interaction, it's essential for robots to accurately perceive transparent objects held by humans and manipulate them safely. This further increases the need for precise and robust depth estimation of transparent objects.

Many attempts have been made to address the problem of transparent object depth estimation by introducing depth restoration models~\cite{sajjan2020clear, chen2023tode, dai2022domain}. The general idea behind those methods is extracting information from the RGB and corrupted depth images to predict the correct depth value of transparent objects. However, restoring the hand-held transparent objects presents many new challenges and a research gap still exists. \textbf{First}, generating data for model training could be much harder in both real-world and simulation settings. At present, there exists no dataset supporting this area of research. \textbf{Second}, hand occlusions bring difficulties in depth restoration. Since objects' appearances are highly related to the hand geometrically, the same object may look different because of hand occlusions, which leaves a higher demand for model generalization ability. \textbf{Additionally}, previous studies on human-to-robot (H2R) handover~\cite{yang2021reactive, christen2023learning} claim their methods are struggled with transparent objects. It remains uncertain whether incorporating depth restoration can improve the overall performance of the transparent object handover.

\begin{figure}[t]
\centerline{\includegraphics[width=0.48\textwidth]{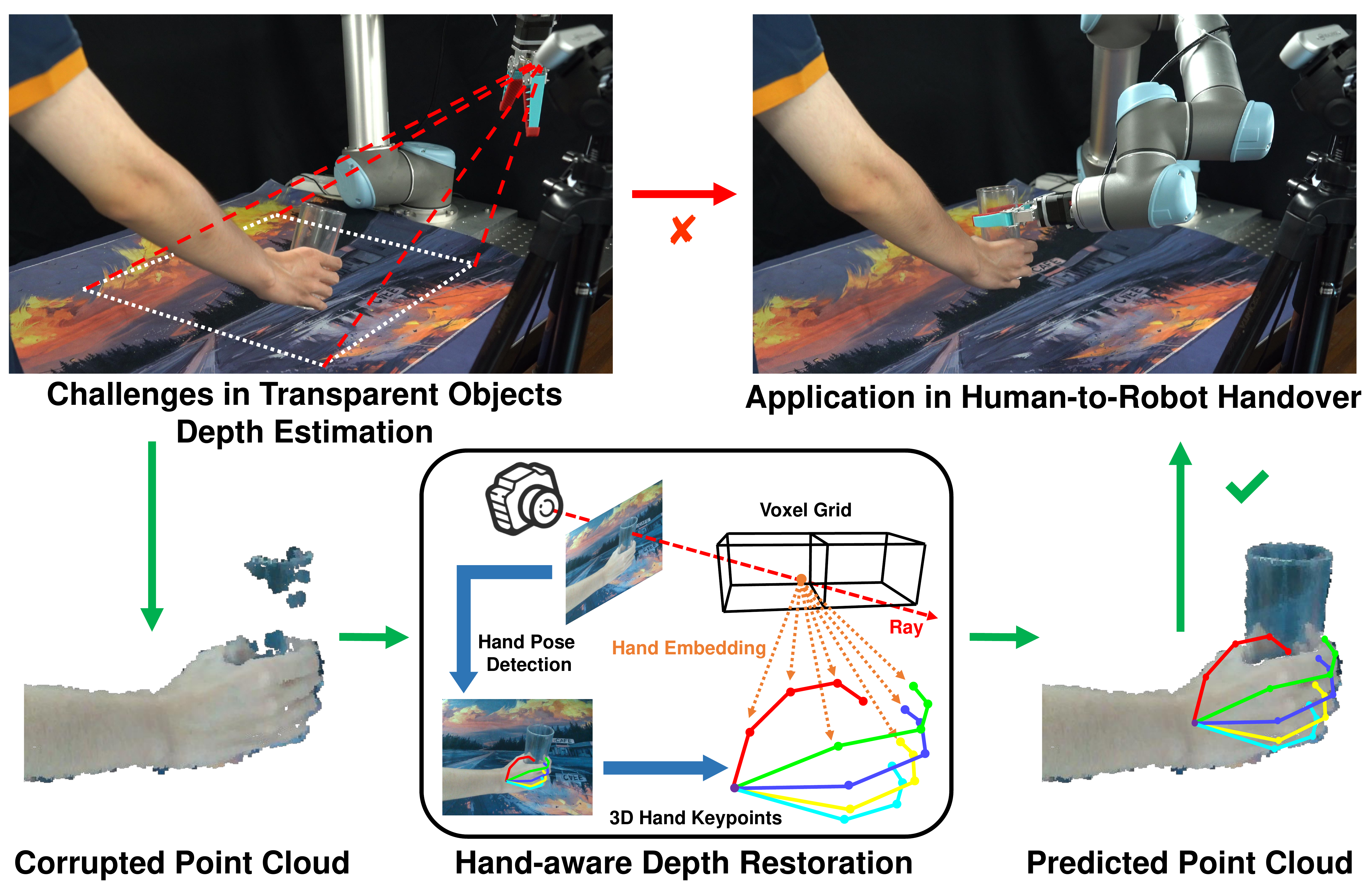}}
\vspace{-7pt}
\caption{RGB-D sensors meet challenges in estimating the depth of transparent objects, especially hand-held objects. We present a hand-aware depth restoration method that reconstructs the corrupted point cloud with the guidance of hand pose information. Our method can be used for human-to-robot handover, highlighting its application value.}
\label{fig: fig1}
\vskip -0.2in
\end{figure}

In this paper, we aim to solve the aforementioned problems from the perspectives of data, algorithm, and application. A synthetic dataset named TransHand-14K is first proposed, which is used for hand-held transparent object perception research. We implement a real-to-sim data generation scheme to ensure the quality and fidelity of data. Based on the dataset TransHand-14K, we present a \textbf{H}and-\textbf{A}ware \textbf{D}epth \textbf{R}estoration (\textbf{HADR}) method for hand-held transparent objects. This method incorporates hand-holding postures to facilitate object reconstruction. Results demonstrate that our method surpasses existing methods, especially in terms of object-level generalization ability. Furthermore, we introduce a human-to-robot handover workflow for transparent objects based on the proposed depth restoration method. This workflow is adaptable to a wide range of commonly encountered transparent objects and is capable of dynamically responding to human movements during the handover process.

In summary, the key contribution of our work includes the following three aspects: 1) a synthetic dataset TransHand-14K for transparent object perception research, 2) a novel depth restoration method HADR for hand-held transparent objects, 3) a human-to-robot transparent objects handover workflow deployed in a real-world system.

\section{Related Work}

\subsection{Depth Restoration of Transparent Objects}
To minimize errors in the depth estimation of transparent objects using 3D sensors, various depth restoration methods are proposed, which utilize RGB and corrupted depth data to reconstruct the objects' shape. Some early works restore the depth with multiple camera viewpoints based on visual hull~\cite{albrecht2013seeing}, stereo-view matching~\cite{phillips2016seeing}, and NeRF~\cite{ichnowski2021dex,dai2023graspnerf}. 
While multi-view methods suffer from relatively long inference time, other works reconstruct transparent objects only using single-view. A global optimization method is introduced in~\cite{sajjan2020clear} that restores corrupted depth by predicting objects' surface normals and edges. In~\cite{zhu2021rgb}, a local implicit neural representation method is proposed. Authors in~\cite{fang2022transcg} design an end-to-end depth completion network with Unet architecture called DFNet. To better extract features from input data, transformer architecture~\cite{vaswani2017attention} is applied to some recent works such as TODE-trans~\cite{chen2023tode} and SwinDRNet~\cite{dai2022domain}. However, all aforementioned methods are designed to reconstruct transparent objects on the tabletop. Depth restoration of hand-held transparent objects remains an open problem, which is crucial for human-robot interaction.

\subsection{Datasets for Transparent Objects Perception}
Training a depth restoration model for transparent objects requires both corrupted and perfect depth data from RGB-D cameras. These data can be generated from either the real world or the simulation environments. A large-scale dataset is proposed in~\cite{sajjan2020clear}, which contains both synthetic data rendering with Blender and real-world data by spraying opaque material on transparent objects. In the work of~\cite{zhu2021rgb}, a synthetic dataset of transparent objects is generated by using the NVIDIA Omniverse platform~\cite{nvidiaOmniverse}. Another large-scale real-world dataset called TransCG is introduced in~\cite{fang2022transcg}, which generates data with an auto-collection pipeline. However, no dataset is currently available for grasped or hand-held transparent objects, as creating such a real-world dataset could require significant time and financial resources. Creating a synthetic dataset also presents challenges, particularly in generating realistic hand poses for objects with various shapes, which is crucial for narrowing the sim-to-real gap.

\section{Methodology}
\subsection{TransHand-14K Dataset Generation}

\begin{figure}[htbp]
\centerline{\includegraphics[width=0.48\textwidth]{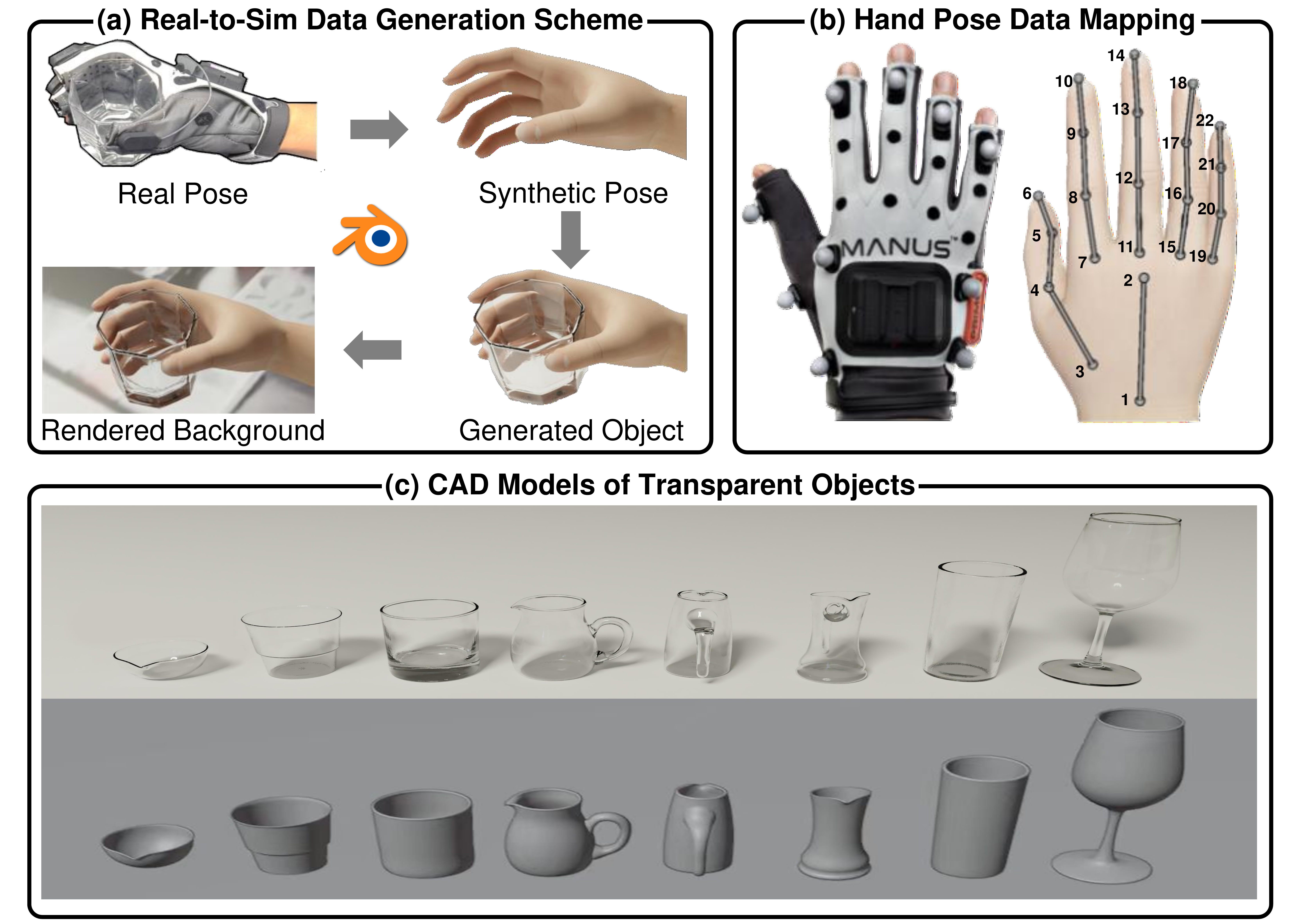}}
\vspace{-7pt}
\caption{TransHand-14K data generation method. (a) A real-to-sim hand pose generation scheme is introduced. (b) We use a MANUS VR glove to capture the pose data. (c) Eight transparent objects are included in TransHand-14K.}
\label{fig: real-to-sim}
\vskip -0.2in
\end{figure}

In this work, we propose a synthetic dataset specifically designed for hand-held transparent objects, named TransHand-14K. Generating such a dataset is more challenging than the conventional object-only datasets~\cite{sajjan2020clear, zhu2021rgb}. Beyond considerations of background setting and lighting, it is also essential to account for the various handhold postures and their realism. Existing works of hand pose generation have some limitations~\cite{miller2004graspit}. For instance, the hand-held object dataset proposed by~\cite{weber2022mixed} only provides six types of grasping schemes for different objects. Additionally, the generated data suffers from problems of unrealistic background and hand poses.

To solve those problems, we propose a real-to-sim data generation scheme. As shown in Fig.~\ref{fig: real-to-sim}(a), we first utilize a MANUS VR glove~\cite{manusmeta2024} (Prime X Haptic) to capture the hand pose data of holding transparent objects, and then map the twin hand poses in Blender~\cite{blender2018}. The synthetic hand in Blender is built on a skeleton model as shown in Fig.~\ref{fig: real-to-sim}(b). To reduce the difficulty of modeling in Blender, we adopt a 22-point skeleton model by creating a virtual point (number 2), which is different from the commonly used 21-point skeleton model. The skeletal control can be achieved by feeding joint angle data captured by the MANUS VR glove. After generating the handhold poses, we manually place the corresponding transparent object into the proper position. 3D scanning is used to model eight common transparent objects with different shapes into Blender for rendering. In addition to ensuring the realism of generated hand posture, the proposed real-to-sim data generation method can easily obtain various grasping poses of the same object, significantly reducing the difficulty of dataset construction.

\vskip -0.1in
\begin{figure}[htbp]
\centerline{\includegraphics[width=0.48\textwidth]{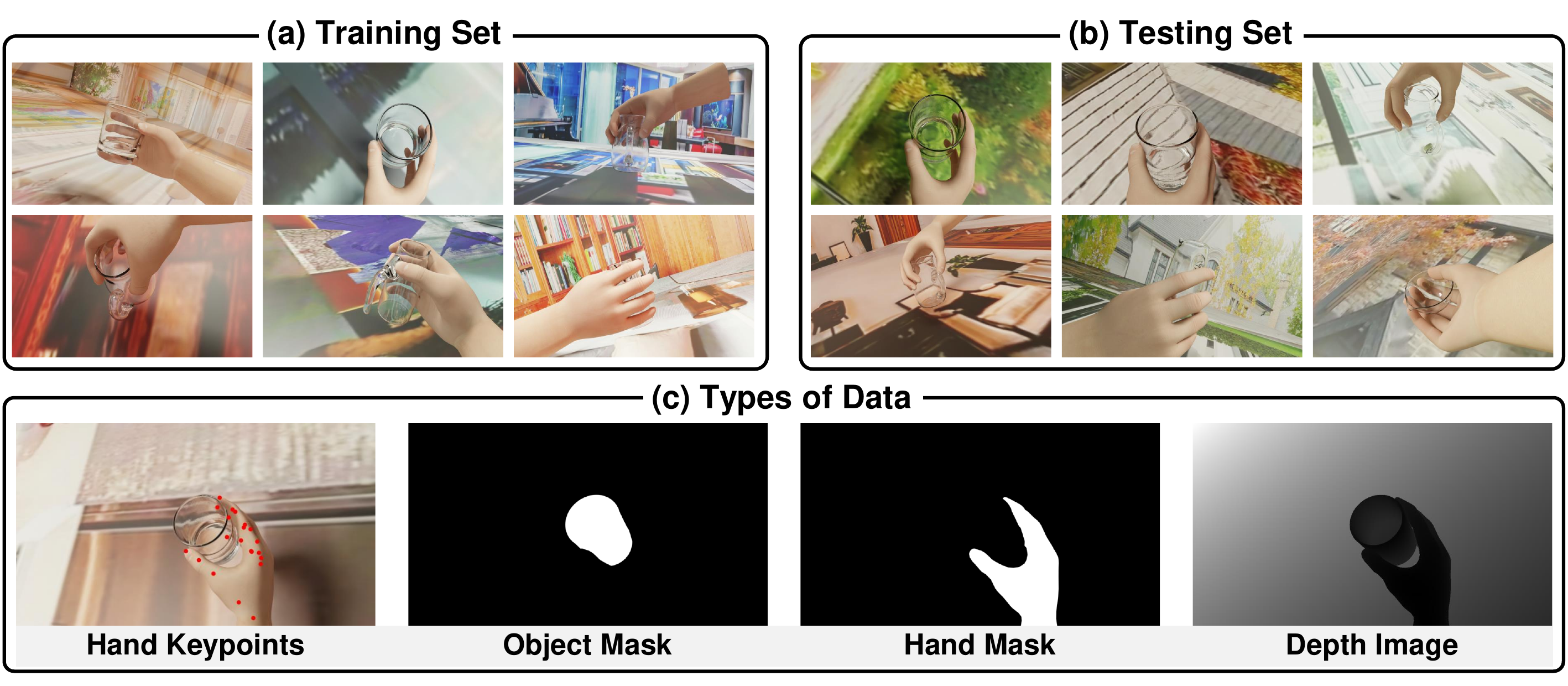}}
\vspace{-7pt}
\caption{Visualization of proposed dataset TransHand-14K.}
\label{fig: dataset visualization}
\end{figure}
\vskip -0.1in

\begin{figure*}[ht]
\centerline{\includegraphics[width=0.98\textwidth]{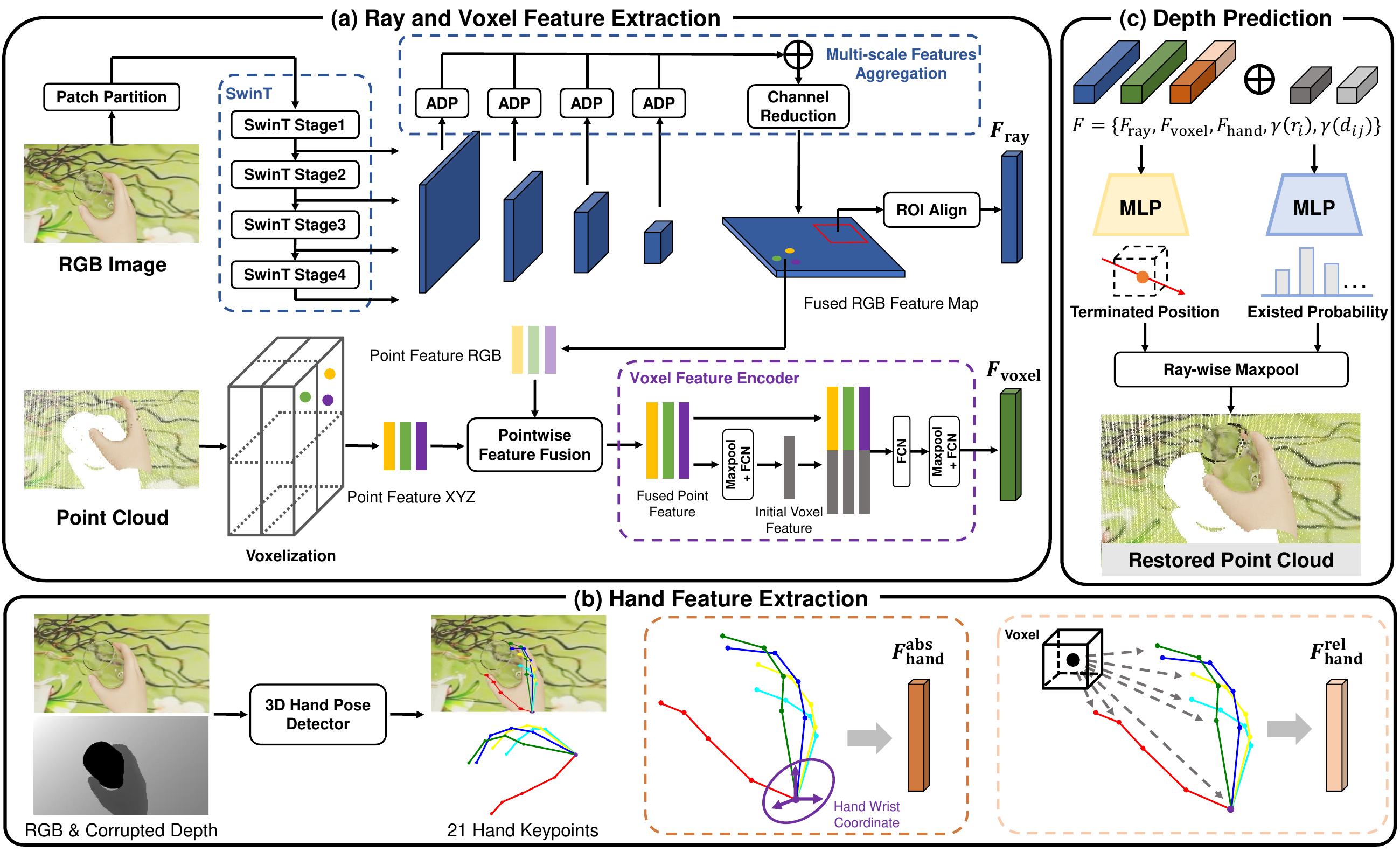}}
\vspace{-7pt}
\caption{Overview of HADR. (a) Ray and voxel features are generated from the RGB and corrupted point cloud. (b) We introduce the handhold pose as an important guidance for depth restoration. This feature is related to geometric and semantic information of hand-object interaction. (c) The terminated probability and position of each ray-voxel pair are predicted. The final restored depth is obtained by ray-wise maxpooling.  }
\label{fig: method}
\vskip -0.2in
\end{figure*}

For the scene and background setting, we follow the work of Trans6D-32K~\cite{yu2023tgf}. Our dataset contains 14100 images with a resolution of 1280 $\times$ 720. As shown in Fig.~\ref{fig: dataset visualization}, TransHand-14K contains RGB image, perfect depth image, objects mask, hand mask, and labels of the hand keypoints. In addition to depth completion, this dataset can be used for other transparent object perception tasks such as detection, recognition, segmentation, and 6D pose estimation.

\subsection{Hand-Aware Depth Restoration}
\textbf{Ray-Voxel Implicit Neural Representation.} We propose a depth restoration method named HADR for hand-held transparent objects. Inspired by the work of~\cite{zhu2021rgb}, our method is based on generating a local implicit neural representation of ray-voxel pairs. The corrupted depth image is first transformed into the point cloud and placed into a voxel grid space in the camera coordinate. A pixel $ I_i $ in the image can be viewed as a camera ray $r_i$ which passes through a set of occupied voxels $\mathbf{V}^{occ}$. The ray-voxel pair~\cite{zhu2021rgb} is defined as $\phi_{ij} = (r_i, v_j)$, where $r_i$ is the camera ray direction for $i$-th pixel and $v_j$ is the intersected voxel with index $j$.

Based on the definition of ray-voxel pair, restoring a missing depth value $ D_i $ is equivalent to finding out it exists in which voxel and the offset distance in this voxel, which can be expressed as
\begin{equation}
    d_{i} = d_{ij'}^{\text{in}} + \delta_{ij'}r_i, \:\: j' = \mathop{\mathrm{argmax}}_{j: v_j^{\text{occ}} \in \mathbf{V}_{r_i}^{\text{occ}}} p_{ij}^{\text{end}},
    \label{equation: depth prediction}
\end{equation}
where $d_i$ is the predicted depth of ray $r_i$; $d_{ij'}^{\text{in}}$ is the entering position of ray-voxel intersected point; $\delta_{ij'}$ is the offset distance;  $p_{ij}^{\text{end}}$ is the probability of ray $r_i$ terminated in the voxel $v_j$. This expression indicates that we can use neural networks to generate an implicit function for predicting the terminated position and probability of each ray-voxel pair, thereby restoring the accurate depth of transparent objects.

\textbf{Ray and Voxel Feature Extraction.} To construct the implicit neural representation, we first generate ray-wise and voxel-wise features for each ray-voxel pair. Fig.~\ref{fig: method}(a) shows the model architecture of extracting these features from the input RGB image and corrupted point cloud.

For the input RGB image $I_C$, we use the swin transformer~\cite{liu2021swin} as the backbone to extract multi-scale image features $\{F^{i}_{\text{rgb}}\}_{i = 1, 2, 3, 4} $, where $ F^{i}_{\text{rgb}} \in \mathbb{R}^{\frac{H}{4i} \times \frac{W}{4i} \times iC}$ and $C=96$ is the embedding dimension of first layer output. We then introduce a features aggregation module to fuse multi-scale 
features into a single dense color feature map $ F^{\text{dense}}_{\text{rgb}} \in \mathbb{R}^{H \times W \times C_{\text{rgb}}}$, where $C_{\text{rgb}}=32$. This is realized by first aligning different features into the same dimension of $\mathbb{R}^{\frac{H}{4} \times \frac{W}{4} \times C}$, and concatenating them together. We then reduce the channels into $C_{\text{rgb}}$ using a $1 \times 1$ convolution layer and conduct bilinear interpolation to resize the feature into $H \times W$. For the ray $r_i$, its corresponding ray feature can be extracted from $F^{\text{dense}}_{\text{rgb}}$ using ROI alignment~\cite{he2017mask}, with an input size of $8 \times 8$ and an output size of $2 \times 2$. The ray feature $F_{\text{ray}}$ is obtained by flattening the output of the ROI alignment.

For the voxel feature, early work~\cite{zhu2021rgb} generates it by sending the position and color of the valid point cloud into a voxel-based PointNet~\cite{yuan2018pcn, qi2017pointnet}. However, this approach may not fully utilize the semantic information provided by the RGB image. To overcome this limitation, we introduce an early-stage point fusion method. We first encode the geometry embedding of point cloud $ P_{\text{xyz}} \in \mathbb{R}^{N \times 16} $ using the relative position of points to their voxel centers. We then project the point cloud with camera intrinsic parameters onto the generated dense color feature map $F^{\text{dense}}_{\text{rgb}}$ and generate point-wise color embedding $P_{\text{rgb}} \in \mathbb{R}^{N \times 16}$ through channel reduction. These two features are concatenated into the fused point feature $P_{\text{fuse}}$. The final voxel feature $F_{\text{voxel}}$ is generated from $P_{\text{fuse}}$ using a voxel feature encoder, which contains a set of voxel-wise maxpooling layers and fully connected layers~\cite{zhou2018voxelnet}.

\textbf{Hand Feature Extraction.} In addition to the ray and voxel feature, HADR incorporates hand pose as important guidance for depth restoration. One of our insights is that the hand pose can provide both geometric and semantic context, enhancing the accuracy and generalization ability of object reconstruction.
As shown in Fig.~\ref{fig: method}(b), we extract the 21 hand keypoints with a pre-trained hand pose detector. In our experiment, a ResNet-18~\cite{he2016deep} is adopted as the backbone to extract the hand feature map from the corrupted depth image, which is then transferred into hand keypoints $X_{\text{hand}} \in \mathbb{R}^{21 \times 3}$ through the weighted average regression~\cite{huang2020awr}.

Two types of hand features are derived from the detected hand keypoints. We first transfer the keypoints into the hand-wrist coordinate and obtain the absolute hand feature $F_{\text{hand}}^{\text{abs}}$. This feature represents the absolute hand-holding pose, which is invariant to the change of camera position and orientation. The second type of hand feature $F_{\text{hand}}^{\text{rel}}$ is the relative position between each hand keypoint and the center of intersected voxel $v_j$. This feature represents the spatial relationship of hand-object interaction. Absolute and relative hand features are flattened and concatenated into the final hand feature $F_{\text{hand}}$.

\textbf{Depth Prediction.} As shown in Fig.~\ref{fig: method}(c), the final embedding feature can be formulated as Equation \ref{equation: final embedding feature}:
\begin{equation}
\label{equation: final embedding feature}
F = F_{\text{ray}} \oplus  F_{\text{voxel}} \oplus F_{\text{hand}} \oplus\gamma(r_i) \oplus \gamma(d_{ij}),
\end{equation}
where $\oplus$ represents the feature-wise concatenation operator, $\gamma(r_i)$ and $\gamma(d_{ij})$ are the ray direction and voxel position embedding, which has shown to be important for the final accuracy~\cite{zhu2021rgb}. Two MLPs are used to decode the feature into the predicted terminated position and the existing probability of each ray-voxel pair. We obtain the final predicted depth value by using ray-wise maxpooling~\cite{zhu2021rgb}, as expressed in Equation \ref{equation: depth prediction}.

\textbf{Training Objective.} The loss function defined in Equation \ref{equation: loss func} is optimized for training:
\begin{equation}
\label{equation: loss func}
L = w_{\text{depth}}L_{\text{depth}} + w_{\text{prob}}L_{\text{prob}} + w_{\text{norm}}L_{\text{norm}},
\end{equation}
where $L_{\text{depth}}$ is the first-order norm between ground truth and predicted depth value; $L_{\text{prob}}$ is the cross entropy loss between the ground truth and predicted voxel existing probability; $L_{\text{norm}}$ is the loss of the surface normal measured by the cosine similarity; $w_{\text{depth}}$, $w_{\text{prob}}$, and $w_{\text{norm}}$ are the weights of different loss components.

\subsection{H2R Handover Workflow of Transparent Objects} \label{section: handover}
\vskip -0.15in
\begin{figure}[h]
\centerline{\includegraphics[width=0.48\textwidth]{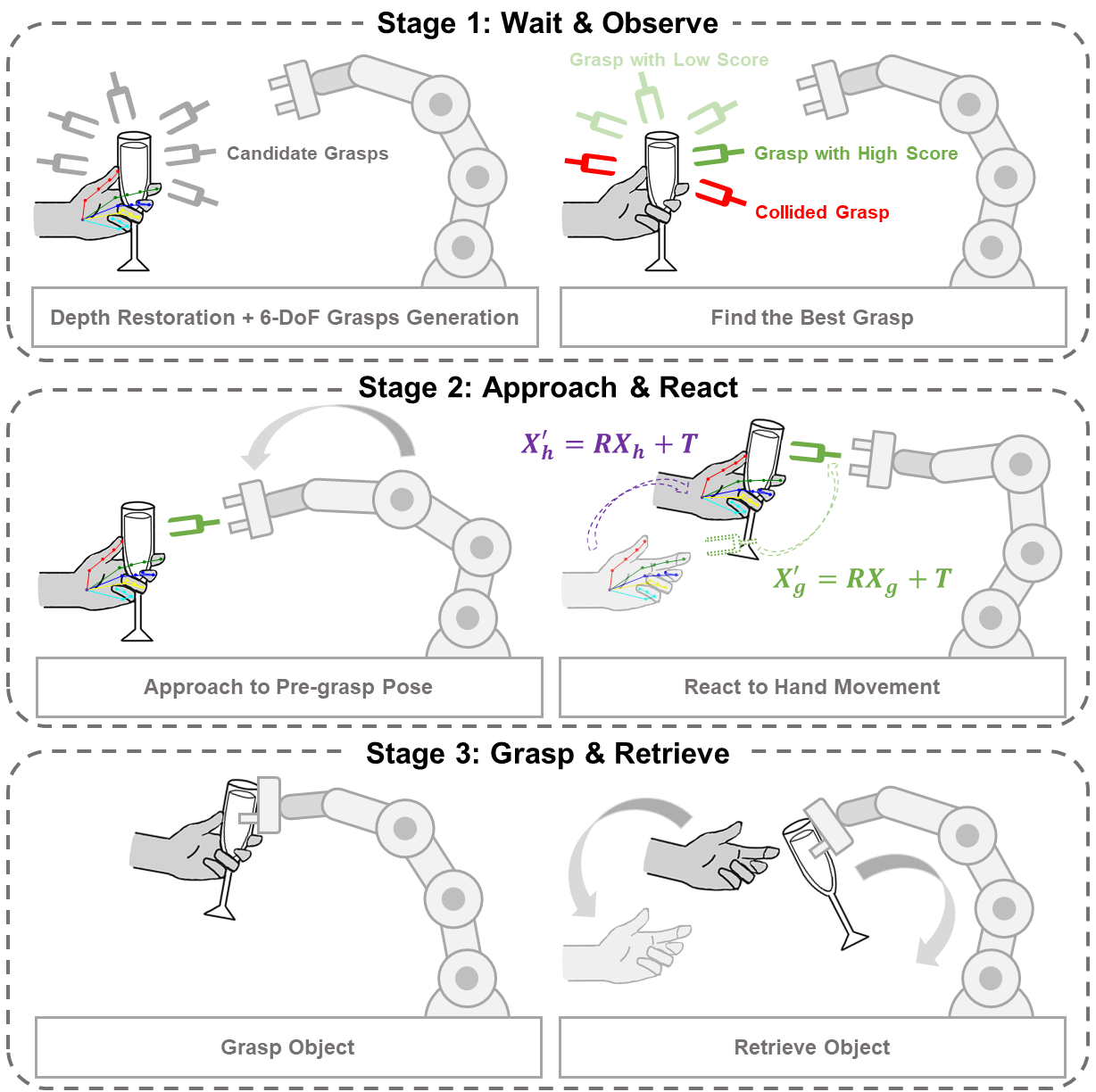}}
\vspace{-7pt}
\caption{Overview of the proposed handover workflow. The whole handover process is divided into three stages: 1)  Wait \& Observe, 2) Approach \& React, and 3) Grasp \& Retrieve.}
\label{fig: handover strategy}
\vskip -0.1in
\end{figure}

Based on the proposed hand-aware depth restoration method, we further develop a handover workflow for transparent objects. As shown in Fig.~\ref{fig: handover strategy}, we divide the whole H2R handover process into three stages: 1) Wait \& Observe, 2) Approach \& React, and 3) Grasp \& Retrieve.

\textbf{Wait \& Observe.} This is the starting phase of the whole handover task. The robot is waiting at the home position, and the depth restoration module is activated to restore the corrupted depth. GraspNet~\cite{mousavian20196} is used to generate a set of 6-DoF grasps $G$ given the restored point cloud of the transparent object. Each potential grasp $g$ is associated with a score $s$. We first select the grasps on the transparent objects based on the generated object mask and then filter the grasps based on the collision detector. For the remaining available grasp, we rescore them based on the metric in Equation \ref{equation: rescore grasp pose}:
\begin{equation}
\label{equation: rescore grasp pose}
C = w_{\text{s}} s + w_{r2h} \mathbf{v}_{\text{r2h}}^\top (R_{\text{g}} \, \mathbf{v}_{\text{r2h}}) + w_{\text{u2d}} \mathbf{v}_{\text{u2d}}^\top (R_{\text{g}} \, \mathbf{v}_{\text{u2d}}),
\end{equation}
where $R_{\text{g}}$ is the rotation of grasp pose, $\mathbf{v}_{\text{r2h}}$ denotes the unit vector pointing from the robot to the human, and $\mathbf{v}_{\text{u2d}}$ denotes the unit vector pointing from the up to the down. We design the metric in this way because a grasp facing toward the user and vertically pointing downward is preferred. $w_s$, $w_{\text{r2h}}$, and $w_{\text{u2d}}$ are the weights, which are set to be $w_\text{s}=1$, $w_{\text{r2h}}=0.4$, and $w_{\text{u2d}}=0.2$. The grasp pose with the highest score will be chosen, and the corresponding pre-grasp pose is defined as 10cm back from the predicted grasp along its z-axis.

\textbf{Approach \& React.} In this stage, the goal of the robot is to approach the pre-grasp pose. Since the human user may move their hand during this stage, the robot should react to the movement in a closed-loop manner. We assume that the absolute holding pose will not be changed significantly during the handover and realize the real-time grasp pose adjustment by tracking the hand keypoints.

We denote the initial and current hand keypoints as $X_{\text{i}}$ and $X_{\text{c}}$ in the dimension of $\mathbb{R}^{21 \times 3}$. The translation vector is computed as $T = X_{\text{c}}^0 - X_{\text{i}}^0$, where $X_{\text{i}}^0$ and $X_{c}^0$ represent the initial and current wrist position. For the rotation, we perform the Singular Value Decomposition on the covariance matrix $H$, defined in Equation \ref{equation: covariance matrix}, 
\begin{equation} \label{equation: covariance matrix}
H = (X_{\text{i}}-X_{\text{i}}^0)^\top (X_{\text{c}}-X_{\text{c}}^0) = U \Sigma V^\top.
\end{equation}
The rotation matrix is then calculated as
$
R = VU^\top,
$
where $R$ should be adjusted by flipping the sign of the last column if its determinant is negative. The translation vector and rotation matrix are then applied to the initial pre-grasp pose to obtain the current one.

\textbf{Grasp \& Retrieve.} Once the robot reaches the pre-grasp position, it will grasp the object in an open-loop manner. We move the final grasp position 5cm forward along the z-axis as we assume the human user will adapt to the robot's grasp motion. Once the gripper closes successfully, the robot will retrieve and drop the object.

\section{Experiments}
\subsection{Depth Restoration Experiments on TransHand-14K}


\textbf{Evaluation Metrics.} We adopt four commonly used metrics for depth estimation~\cite{sajjan2020clear}. 1) \textbf{RMSE:} the root mean square error between the predicted result and ground truth. 2) \textbf{REL:} the mean value of absolute relative difference. 3) \textbf{MAE:} the mean absolute error between the predicted result and ground truth. 4) \textbf{Threshold $\delta$:} the percentage of data smaller than the given threshold $\delta$. $ \max \left( d_i / d_i^*, d_i^* / d_i \right) < \delta$, where $d$ represents the predicted depth, $d^*$ represents the ground truth depth, and $\delta \in \{1.05, 1.10, 1.25\}$. All metrics are evaluated on the transparent objects area.

\textbf{Implementation Details.} We set the image resolution to be 224 $\times$ 224 for both RGB and depth input. The RGB input is further augmented by adjusting the color space and incorporating additional elements such as bright patches, blur, and noise. We train the network for 100 epochs and choose Adam as the optimizer with a learning rate of 1e-3 for the first 80 epochs and 1e-4 for the last 20 epochs. The number of voxels used for restoration is $8^3$. We set $w_{\text{depth}}=200$, $w_{\text{prob}}=10$, and $w_{\text{norm}}=0.5$ in training.

\textbf{Comparison with SOTA.} We compare the proposed HADR with several state-of-the-art methods, including TODE-Trans~\cite{chen2023tode}, TransCG~\cite{fang2022transcg}, SwinDRNet~\cite{dai2022domain}, and LIDF \cite{zhu2021rgb}. All baselines are trained on the proposed dataset TransHand-14K. We split the training, validation, and testing set according to the ratio of 7:2:1.

Table~\ref{table: compare with sota} reports the comparison results. We first compare methods on the known category objects, where models are trained and evaluated on all eight different shapes of transparent objects. Our method achieves the best performance in terms of RMSE, MAE, $\delta_{1.05}$, and $\delta_{1.10}$. For the other two metrics, our method is slightly weaker than SwinDRNet. For the unknown category evaluation, we choose six objects for training and the other two for evaluation. Our method significantly outperforms other methods in terms of all metrics, demonstrating the strong generalization ability of the model. This better performance can be attributed to our hand-aware design, which enables the model to effectively transfer similar hand-object interaction patterns to unknown objects. Additionally, unlike LIDF \cite{zhu2021rgb}, our method can directly produce high-quality final results without needing a depth refinement process.


\begin{figure}[t]
\vskip 0.05in
\centerline{\includegraphics[width=0.46\textwidth]{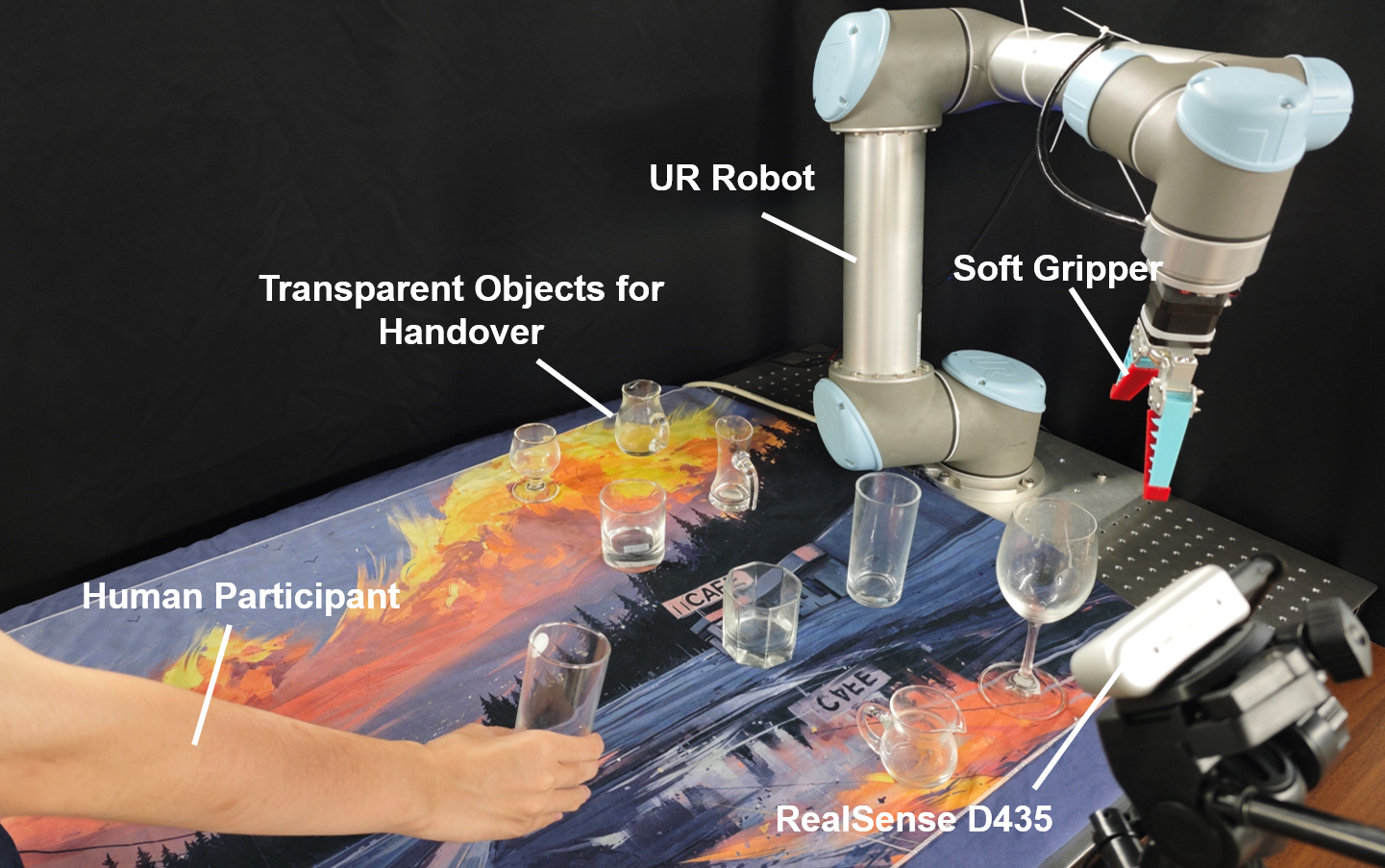}}
\vspace{-7pt}
\caption{The setup of our real-world robot experiment.}
\label{fig: real robot setup}
\vskip -0.25in
\end{figure}

\begin{figure*}[t]
\centerline{\includegraphics[width=0.97\textwidth]{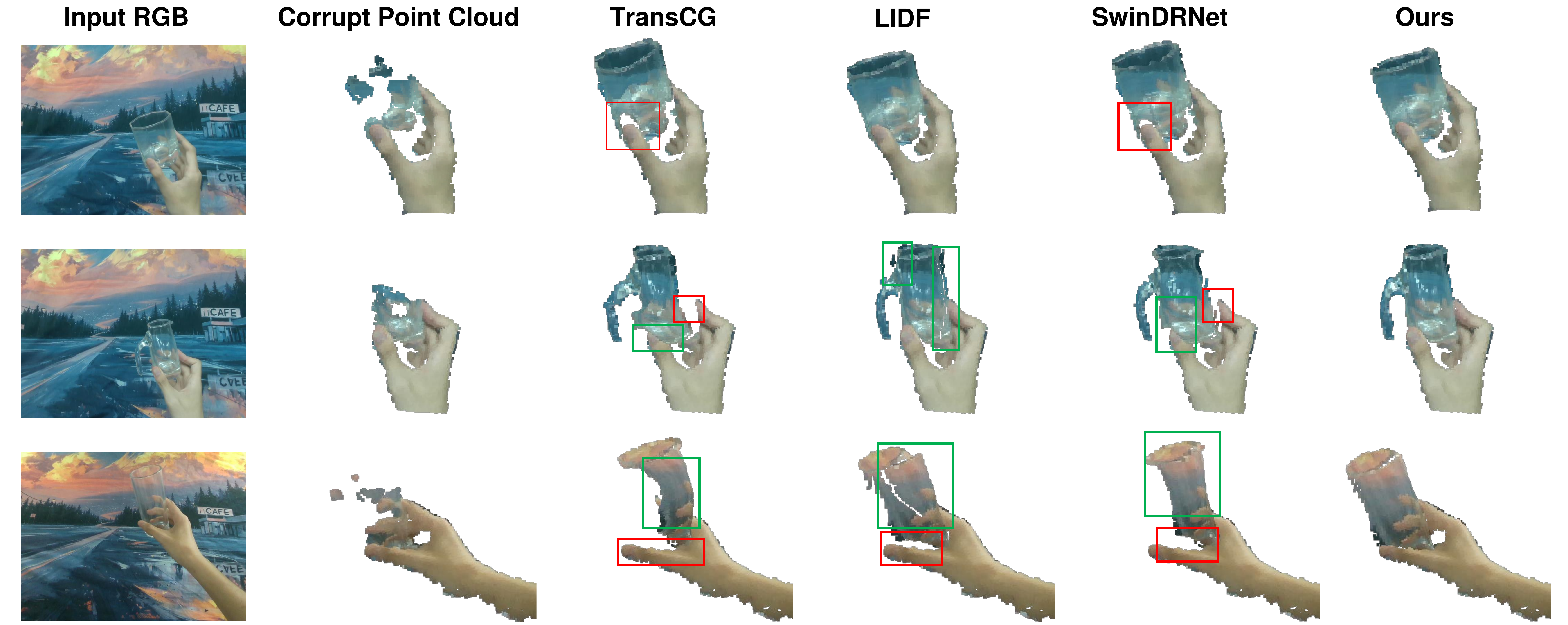}}
\vspace{-7pt}
\caption{Qualitative evaluation in real scenarios. We compare the proposed HADR to state-of-the-art methods. Defects including deformation (in the color of green) and spatial shifting (in the color of red) are highlighted. For better visualization, we remove point cloud of the background.}
\label{fig: qualitative results}
\vskip -0.15in
\end{figure*}

\vskip -0.05in
\begin{table}[htpb]
\caption{Quantitative comparison with state-of-the-art methods. }
\vskip -0.02in
\label{table: compare with sota}
\vspace{-19pt} 
\begin{center}
\resizebox{\columnwidth}{!}{%
\begin{tabular}{@{}l|cccccc@{}}
\toprule
\textbf{Methods}     & \textbf{RMSE}$\downarrow$ & \textbf{REL}$\downarrow$ & \textbf{MAE}$\downarrow$ & $\boldsymbol{\delta_{1.05}}\uparrow$ & $\boldsymbol{\delta_{1.10}}\uparrow$ & $\boldsymbol{\delta_{1.25}}\uparrow$ \\
\midrule
\multicolumn{7}{c}{Known Category Evaluation} \\
\midrule
TODE-Trans \cite{chen2023tode}  & 0.024 & 0.056 & 0.017 & 57.03 & 82.55 & 98.61 \\
TransCG \cite{fang2022transcg}     & 0.012 & 0.026 & 0.008 & 86.91 & 96.96 & 99.76 \\
SwinDRNet \cite{dai2022domain}   & \textbf{0.009} & \textbf{0.015} & \textbf{0.005} & 94.82 & 98.26 & \textbf{99.87} \\
LIDF+Refine \cite{zhu2021rgb}     & 0.014 & 0.029 & 0.010 & 85.18 & 96.08 & 99.60 \\
HADR (Ours)   & \textbf{0.009} & 0.016 & \textbf{0.005} & \textbf{95.22} & \textbf{98.44} & 99.81 \\
\midrule
\multicolumn{7}{c}{Unknown Category Evaluation} \\
\midrule
TODE-Trans \cite{chen2023tode}   & 0.052 & 0.055 & 0.037 & 62.86 & 89.26 & 97.21 \\
TransCG  \cite{fang2022transcg}     & 0.027 & 0.064 & 0.020 & 51.54 & 78.54 & 98.79 \\
SwinDRNet \cite{dai2022domain}   & 0.022 & 0.049 & 0.015 & 65.07 & 86.79 & 98.79 \\
LIDF+Refine \cite{zhu2021rgb}       & 0.026 & 0.063 & 0.021 & 52.66 & 78.69 & 97.90 \\
HADR (Ours)   & \textbf{0.018} & \textbf{0.042} & \textbf{0.014} & \textbf{72.74} & \textbf{90.06} & \textbf{99.05} \\

\bottomrule
\end{tabular}
}
\end{center}
\vskip -0.3in
\end{table}

\begin{table}[htpb]
\caption{Ablation Study of Features and Modules.}
\vskip -0.02in
\label{table: ablation test}
\vspace{-17pt} 
\begin{center}
\resizebox{\columnwidth}{!}{%
\begin{tabular}{@{}l|cccccc@{}}
\toprule
& \textbf{RMSE}$\downarrow$ & \textbf{REL}$\downarrow$ & \textbf{MAE}$\downarrow$ & $\boldsymbol{\delta_{1.05}}\uparrow$ & $\boldsymbol{\delta_{1.10}}\uparrow$ & $\boldsymbol{\delta_{1.25}}\uparrow$ \\
\midrule
w/o Hand Feature      & 0.012 & 0.019 & 0.006 & 93.35 & 97.79 & 99.70 \\
w/o MS Aggregation  & 0.011 & 0.020 & 0.007 & 92.25 & 97.49 & 99.70 \\
w/o Point Fusion  & \textbf{0.009} & 0.017 & 0.006 & 94.40 & 98.37 & 99.80 \\
2D Hand Feature      & 0.010 & 0.017 & 0.006 & 94.09 & 98.30 & 99.77 \\
Full Model  & \textbf{0.009} & \textbf{0.016} & \textbf{0.005} & \textbf{95.22} & \textbf{98.44} & \textbf{99.81} \\
\bottomrule
\end{tabular}
}
\end{center}
\vskip -0.1in
\end{table}

\textbf{Ablation Study.} We conduct ablation experiments to analyze the different components of the proposed method. To assess the impact of hand-ware on performance, we compare the model with two other versions: the model excludes the hand pose feature and uses 2D hand keypoints as the feature. As shown in Table~\ref{table: ablation test}, the hand pose feature contributes to the depth restoration result. Although the 2D keypoints feature produces a slightly inferior performance compared with the 3D version, in real-world applications, obtaining the 2D keypoints could be relatively easy and robust in some situations. We also evaluate the effectiveness of multi-scale feature aggregation and point-wise feature fusion. Results indicate that removing these two modules will lead to a drop in overall performance.

\subsection{Real-World Experiments}
\vskip -0.05in
We conduct real-world experiments to verify the effectiveness of HADR and handover workflow.

\textbf{Experiment Setup.} Fig.~\ref{fig: real robot setup} illustrates the setting of our real robot system. A UR5 robot arm equipped with a soft gripper is rigidly mounted on the table, which is used to perform the robot handover task. For perception, we employ a RealSense D435 camera with an RGB-D stream. The camera is placed in a third-person view with extrinsic calibrated. A human participant stands on the opposite side of the table, holding a transparent object in hand and passing it to the robot. We choose the tablecloth with intricate patterns and rich colors as the background to examine the robustness of our method.

\textbf{Implementation.} In addition to implementing the proposed depth restoration method, some other perception models are used in the real-world experiment. To obtain the segmented area of transparent objects, we train a segmentation model in the Unet architecture with a ResNet-34~\cite{he2016deep} as the encoder using the TransHand-14K dataset. For the hand pose detector, we find the weighted average regression from the depth image is vulnerable to occlusions and noise. Instead, we first generate 2.5D hand keypoints using an off-the-shelf RGB-based hand pose detector~\cite{lugaresi2019mediapipe}, and then extend it into 3D keypoints using the depth value of the wrist keypoint position. The pre-trained GraspNet-baseline~\cite{fang2020graspnet} is used to generate 6-DoF grasping poses. All the aforementioned models are implemented using a computer with an NVIDIA 3080 GPU. In the experiment, the whole perception pipeline achieves an inference speed of 11 FPS, allowing real-time depth restoration.

\textbf{Qualitative Evaluation in Real Scenarios.} To better illustrate the sim-to-real performance of the proposed method, a qualitative evaluation is conducted by comparing it to the state-of-the-art methods in real scenarios. As shown in Fig.~\ref{fig: qualitative results}, it can be observed that the reconstructed object point cloud exists deformation (highlighted with green boxes) and spatial shifting (highlighted with red boxes). Results demonstrate that our method reconstructs transparent objects with greater accuracy and robustness in real-world settings.

\textbf{Handover Experiment.} We conduct a human-to-robot handover experiment using the proposed handover workflow. In the experiment, eight transparent objects with different shapes are used. To better illustrate the generalization ability of our model, three of them are in novel shapes. For each object, we perform six independent trials with different ways of holding. The baseline chosen for comparison is GraspNet without depth restoration. In addition to assessing the overall success rate, we also compare the number of objects that exceed the specific success rate threshold  $\delta \in \{0.5, 0.8, 1.0\}$.

\vskip -0.05in
\begin{table}[htpb]
\caption{Results of handover experiment.}
\label{table: handover}
\vspace{-18pt} 
\begin{center}
\resizebox{\columnwidth}{!}{%
\begin{tabular}{@{}cccccc@{}}
\toprule
\textbf{Methods} & \textbf{Success Attempts} & \textbf{Success Rate} & $\boldsymbol{\delta_{0.5}\uparrow}$ & $\boldsymbol{\delta_{0.8}\uparrow}$ & $\boldsymbol{\delta_{1.0}\uparrow}$\\
\midrule
GraspNet & 5/48  & 10.4\% & 0/8 & 0/8 & 0/8\\
HADR + GraspNet & 34/48  & 70.8\% & 6/8 & 3/8 & 1/8 \\
\bottomrule
\end{tabular}
}
\end{center}
\vskip -0.2in
\end{table}

Table \ref{table: handover} reports the experimental results. Without depth restoration, GraspNet struggles to generate the correct grasp pose for handover due to the incomplete and inaccurate depth value. The proposed depth restoration method can significantly improve the handover success rate, from 10.4\% to 70.8\%. Six among eight objects achieve a success rate of over 50\%.

\vskip -0.15in
\section{Conclusion}

In this study, we introduce a novel solution for restoring the depth of hand-held transparent objects and highlight its great value in human-robot interaction. We present TransHand-14K, a synthetic dataset generated through a real-to-sim data generation scheme, effectively bridging the gap between simulated and real-world environments. Our proposed method, HADR, leverages hand posture information to enhance depth restoration, achieving superior reconstruction results and better generalization capabilities compared to existing methods. Additionally, we develop a handover workflow that leverages our depth restoration technique. Evaluations on both depth restoration and human-to-robot handover validate the effectiveness of our approach.

\bibliographystyle{IEEEtran}
\bibliography{ref}

\end{document}